\documentclass[conference]{IEEEtran}
\IEEEoverridecommandlockouts
\usepackage{cite}
\usepackage{amsmath,amssymb,amsfonts}
\usepackage{algorithmic}
\usepackage{graphicx}
\usepackage{textcomp}
\usepackage{xcolor}
\def\BibTeX{{\rm B\kern-.05em{\sc i\kern-.025em b}\kern-.08em
    T\kern-.1667em\lower.7ex\hbox{E}\kern-.125emX}}
\usepackage{siunitx}

\begin{document}

\title{A Hybrid Hierarchical 1D-CNN-BiLSTM Framework for Extractive Summarization of Biomedical and Clinical Text}

\author{\IEEEauthorblockN{Saad Bin Ather}
\IEEEauthorblockA{\textit{CS Department} \\
\textit{NUCES-FAST}\\
Lahore, Pakistan \\
l230838@lhr.nu.edu.pk }
\and
\IEEEauthorblockN{Muhammad Saif}
\IEEEauthorblockA{\textit{CS Department} \\
\textit{NUCES-FAST}\\
Lahore, Pakistan \\
l230969@lhr.nu.edu.pk }
\and
\IEEEauthorblockN{Ali Hassan Khan}
\IEEEauthorblockA{\textit{CS Department} \\
\textit{NUCES-FAST}\\
Lahore, Pakistan \\
l230924@lhr.nu.edu.pk }
\and
\IEEEauthorblockN{Manzer Abbas}
\IEEEauthorblockA{\textit{CS Department} \\
\textit{NUCES-FAST}\\
Lahore, Pakistan \\
l230970@lhr.nu.edu.pk }

\and
\IEEEauthorblockN{Hajra Waheed}
\IEEEauthorblockA{\textit{CS Department} \\
\textit{NUCES-FAST}\\
Lahore, Pakistan \\
hajra.waheed@nu.edu.pk}

}

\IEEEpubid{\makebox[\columnwidth]{%
979-8-3195-0206-3/26/\$31.00~\copyright~2026 IEEE
\hfill}\hspace{\columnsep}\makebox[\columnwidth]{}}

\maketitle

\IEEEpubidadjcol

\begin{abstract}
Large language models have made abstractive summarization remarkably fluent, but fluency has come at a cost. Models trained to generate freely over long documents may hallucinate, producing summaries that read well but drift from what the source actually says. This risk is especially serious in biomedical and clinical text, where fabricated details can have serious consequences. We address this by removing free-form generation from the pipeline rather than trying to constrain it. Our Hybrid Hierarchical CNN-LSTM Summarizer reframes summarization as extractive sentence selection: a stacked, multi-kernel convolutional network progressively refines contextual representations, while a bidirectional LSTM models long-range dependencies across the document. A lightweight scoring head predicts sentence importance using binary cross-entropy supervision. For PubMed, a document-adaptive mean-plus-standard-deviation threshold with a Top-3 fallback determines extraction; clinical evaluations instead use a fixed top-30\% sentence-selection policy. Because output sentences are copied verbatim, the framework prevents source-external generation, although omissions and contextual inaccuracies remain possible. Evaluated on PubMed and transferred without clinical fine-tuning to MIMIC-CXR and MIMIC-IV BHC, the model demonstrates competitive extractive summarization performance, with threshold and component ablations
supporting adaptive extraction and the
combined CNN--BiLSTM architecture. Clinical factuality analysis further highlights limitations in preserving important findings and negations. These results suggest that structural constraints offer a promising direction for source-grounded biomedical summarization, while emphasizing the importance of clinical completeness and contextual fidelity.
\end{abstract}

\begin{IEEEkeywords}
Extractive Text Summarization, Clinical Natural Language Processing, Hierarchical 1D-CNN, Bidirectional LSTM, Electronic Health Records (EHR), Factuality Preservation, Biomedical Document Extraction.
\end{IEEEkeywords}

\section{Introduction}
Text summarization is broadly categorized into two main approaches: extractive and abstractive. 
Extractive summarization identifies and selects key
sentences directly from the source text, avoiding
source-external generation. However, selected sentences
may omit important context or form disconnected
fragments rather than coherent prose
\cite{azam2025current}. Abstractive methods, especially those built on modern large language models, instead read the source and generate an entirely new summary, producing fluent, human-like text that extractive methods cannot match \cite{zayed2025automatic}.

That fluency comes at a cost. When attention spreads thin across long inputs, small inconsistencies compound: the model does not fail by refusing to answer, but by answering confidently with summaries that are fluent, plausible, and factually wrong. This behavior, commonly called hallucination, is especially dangerous in domains such as biomedical and clinical text, where an invented detail is not a stylistic imperfection but a factual error with real consequences \cite{zayed2025automatic}.

Much of the recent literature manages hallucination through
decoding constraints, factuality scoring, or post-hoc
verification. We take a different position: rather than
generating a summary and subsequently verifying it, we
investigate extractive selection as a source-grounded
alternative. Prior work has demonstrated the effectiveness
of pretrained embeddings with recurrent sentence classifiers
\cite{biswas2025efficient}, CNN-LSTM hybrids for biomedical
summarization \cite{bedi2024extractive}, and hierarchical
document modeling \cite{wang2024study}. Unlike conventional
parallel multi-kernel CNNs, our architecture applies
convolutional layers sequentially, each operating on
the preceding layer's feature map before bidirectional
contextual modeling. We further investigate adaptive
sentence selection through a document-specific
mean-plus-standard-deviation threshold.

We address this gap with a Hybrid Hierarchical CNN-LSTM Summarizer that reframes summarization as sentence selection rather than text generation, preventing source-external generation while retaining the possibility of contextual omissions. The primary contributions of this work are as follows:
\begin{enumerate}
    \item \textbf{Hierarchical Architecture:} A sequential 1D-CNN pipeline that progressively refines local sentence features before modeling global document context via a BiLSTM.
    \item \textbf{Adaptive Extraction:} A dynamic thresholding strategy that adapts to the unique score distribution of each document, eliminating the need for rigid, fixed-length summaries.
    \item \textbf{Cross-Domain Evaluation:} Evaluation across PubMed, MIMIC-CXR, and MIMIC-IV BHC, including threshold ablation, comparative clinical evaluation, and automated factuality assessment.
\end{enumerate}

\section{Related Work}

\subsection{Surveys of Extractive Summarization}
Extractive summarization has been studied from many angles, and two recent reviews give a useful sense of where the field currently stands. Azam et al. \cite{azam2025current} surveyed 145 papers and proposed a layered generic architecture for extractive systems, moving through preprocessing, feature extraction, sentence scoring, base modeling, sentence selection, and post-processing. Their review found that deep learning methods, including CNN and LSTM hybrids, now dominate across news, scientific, and social media domains. Zayed et al. \cite{zayed2025automatic} took a broader view, comparing statistical, graph-based, and neural approaches across both extractive and abstractive summarization, specifically flagging hallucination, prompt sensitivity, and the lack of standardized cross-domain benchmarks as open problems. This directly motivates our work, since limiting hallucination through structural constraints on generation, rather than correcting it after the fact, is the central rationale for our purely extractive approach.

\subsection{Recurrent and Hybrid CNN-LSTM Architectures}
At the model level, pairing pretrained embeddings with recurrent networks has proven highly effective for extractive tasks. Framing extraction as binary sentence classification, a simple LSTM paired with BERT embeddings was shown to outperform baseline classifiers on the Cornell Newsroom dataset \cite{biswas2025efficient}. Comparative studies reinforce this; LSTMs and GRUs have been found to perform at a similarly high level for scientific journal summarization, decisively beating classical latent semantic analysis on ROUGE scores \cite{fitrianah2022extractive}.

Building on these recurrent foundations, recent work has explored combining LSTMs with CNNs. Bedi et al. \cite{bedi2024extractive} proposed a Deep Dense LSTM network followed by a CNN layer for biomedical transcript summarization, demonstrating strong transferability to medical text. 
Transformer-based approaches have also demonstrated
the effectiveness of contextual representations
for biomedical summarization
\cite{datta2024extractive}. Our framework instead
investigates sequential convolutional refinement
combined with bidirectional recurrent modeling
as an alternative architectural design.

Beyond sentence-wise classification, Zhong et al.
\cite{zhong2020matchsum} introduced MatchSum, which
formulates extractive summarization as semantic matching
between documents and candidate summaries. Unlike
this summary-level matching approach, our framework
models contextual representations through sequential
convolutions and a BiLSTM before sentence selection.

\subsection{Hierarchical Structure and Pipeline Complexity}
Handling long, structurally complex documents has pushed the field beyond single flat encoders. Explicitly modeling both local topics and overarching document hierarchy has been shown to outperform flat baselines like BERTSUM on lengthy PubMed and ArXiv papers \cite{wang2024study}. In multi-document settings, classical feature scoring has been combined with a CNN-RNN module to reduce redundancy \cite{purushothaman2026enhancing}.

Recent clinical summarization research has increasingly
explored large language models and factuality-aware
generation. Van Veen et al. \cite{vanveen2024clinical}
investigated adapted LLMs across multiple clinical
summarization tasks, while Hegselmann et al.
\cite{hegselmann2024faithful} examined how training-data
quality affects the faithfulness of generated patient
summaries. These approaches offer flexible generation,
whereas our extractive framework constrains outputs
to source text, trading generative flexibility for
source-grounded selection.

These studies demonstrate the value of recurrent,
convolutional, hierarchical, and generative approaches
to summarization. Our work investigates a distinct
architectural combination: sequentially stacked
convolutions, rather than parallel multi-kernel
branches, followed by bidirectional contextual modeling
and document-adaptive extraction. The objective is not
to eliminate all factual errors, but to prevent
source-external generation while evaluating the
trade-offs in summary completeness and clinical fidelity.

\section{Methodology}

\begin{figure}[htbp]
\centering
\includegraphics[width=\columnwidth]{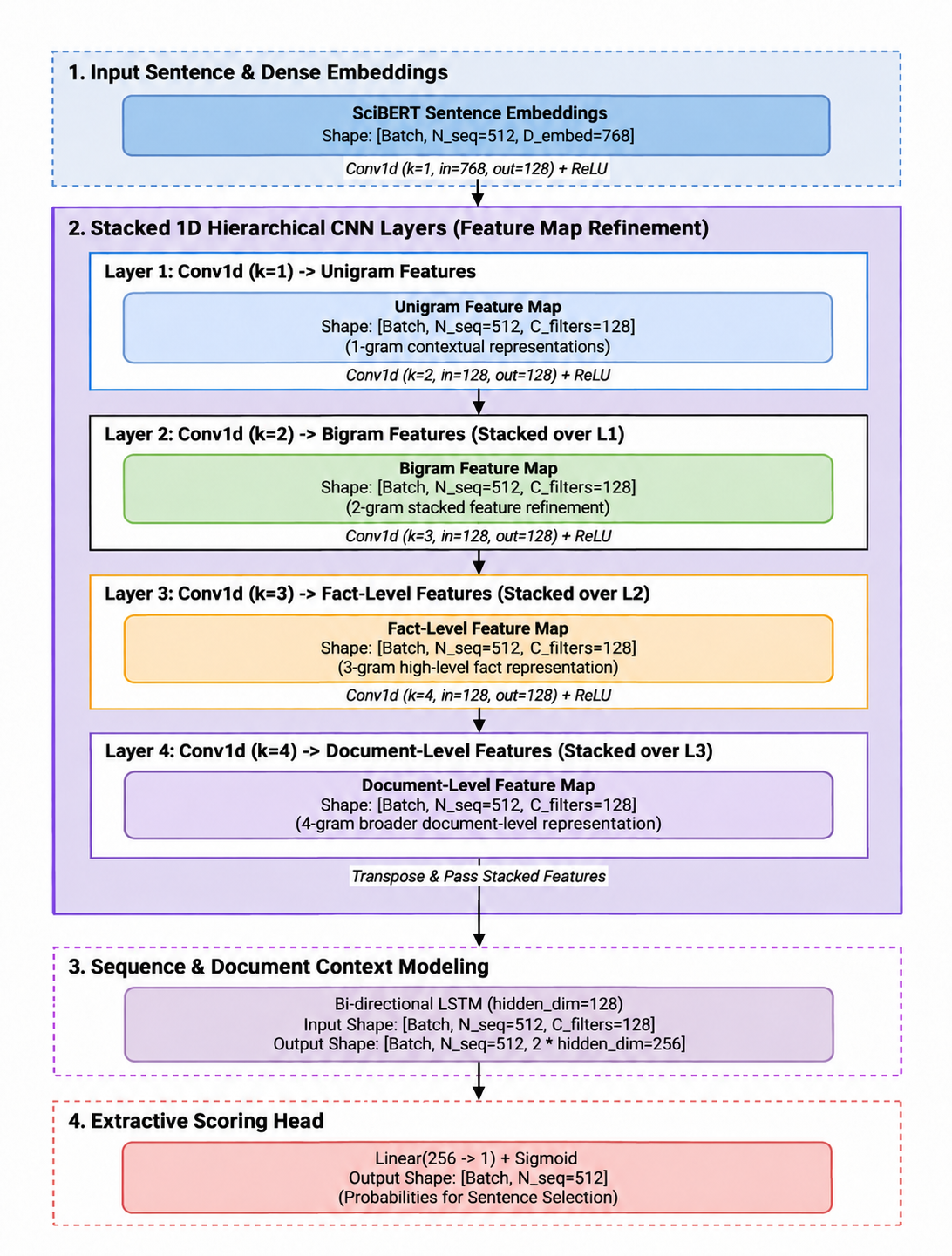}

\caption{Proposed four-layer sequential
1D-CNN--BiLSTM architecture. Each input
position represents a SciBERT sentence
embedding. Sequential convolutional layers
refine inter-sentence features before
bidirectional contextual modeling and
sentence-level importance scoring.}

\label{fig:architecture}
\end{figure}

This research proposes a Hybrid Hierarchical Stacked 1D-CNN--BiLSTM Extractive Summarizer. In contrast to conventional multi-kernel convolutional architectures, which apply several kernel widths in parallel and concatenate their outputs, our design adopts a strictly sequential, stacked hierarchy: each convolutional layer operates directly on the feature map produced by the layer before it, allowing local sentence-context representations to be progressively refined into broader contextual features before contextualization and scoring. The complete pipeline, including tensor dimensions at each stage, is illustrated in Fig.~\ref{fig:architecture}.

\subsection{Sentence Segmentation and Dense Representations}

Each biomedical document is represented as a
sequence of sentences. SciBERT
(\texttt{allenai/scibert\_scivocab\_uncased})
encodes each sentence into a 768-dimensional
contextual representation, using its [CLS]
embedding. The sentence representations are
stacked in their original document order
and padded or truncated to 512 sentence
positions, yielding

\begin{equation}
\mathbf{X}^{(0)} \in
\mathbb{R}^{B \times 512 \times 768},
\label{eq:embedding}
\end{equation}

where $B$ denotes batch size. The tensor is
transposed to $\mathbb{R}^{B\times768\times512}$
before convolution. The same sentence-level
representation principle is used for
clinical documents, which are segmented
using NLTK prior to encoding.

\subsection{Stacked Hierarchical 1D Convolutional Feature Refinement}
The central architectural contribution is a sequentially stacked 1D convolutional network where each kernel operates on the preceding layer's output (Fig.~\ref{fig:architecture}). This design encourages progressive compositional abstraction rather than aggregating parallel fixed-width views of the same input.


Formally, each layer $l \in \{1,2,3,4\}$ applies a 1D convolution with kernel width $k_l = l$ directly to the output of the preceding layer:
\begin{equation}
\resizebox{0.89\columnwidth}{!}{%
$\mathbf{X}^{(l)} = \text{ReLU}\left(\text{Conv1D}_{k=l}\left(\mathbf{X}^{(l-1)}\right)\right) \in \mathbb{R}^{B \times 128 \times 512}, \quad l = 1,\dots,4.$%
}
\label{eq:cnnstack}
\end{equation}
All convolutional layers employ 128 filters with
\texttt{padding='same'}, preserving the sequence
length across the stack. Kernel widths
$k\in\{1,2,3,4\}$ are applied sequentially,
with each layer refining the feature map
produced by its predecessor. 
Each sequence position corresponds to
one sentence embedding. Consequently,
the convolutional stack captures
increasingly broad inter-sentence context
through sequential kernels of widths
1, 2, 3, and 4. Each successive layer
refines features produced by the preceding
layer rather than operating independently
on the original embeddings.

\subsection{Sequential Contextual Modeling via BiLSTM}
The final convolutional feature map
$\mathbf{X}^{(4)}$ is transposed to
$\mathbf{H}_{\mathrm{cnn}}\in
\mathbb{R}^{B\times512\times128}$ and passed
through a single-layer bidirectional LSTM
with 128 hidden units per direction.
Concatenating the forward and backward states
produces contextual representations
$\mathbf{H}_{\mathrm{lstm}}\in
\mathbb{R}^{B\times512\times256}$.
The contextualized positions follow the
dataset-specific encoding procedures
described in Section III-A.
\subsection{Extractive Scoring Head and Training Objective}
Each contextualized position receives a score
through a linear projection and sigmoid:
$\hat{y}_i=\sigma(\mathbf{W}_c
\mathbf{H}_{\mathrm{lstm},i}+b_c)$.
Training uses masked binary cross-entropy
against binary selection labels, excluding
padded positions. 
Each valid sentence position corresponds
to one sentence embedding, one binary
oracle selection label, and one predicted
saliency score. Oracle labels indicate
whether the corresponding source sentence
belongs to the extractive reference.
During inference, selected indices map
directly to their original source sentences,
which are concatenated in document order
without generating new text.

\subsection{Adaptive Sentence Selection}
Rather than applying a fixed top-$k$ cutoff,
we investigate document-adaptive sentence
selection using a score-distribution threshold:

\begin{equation}
\begin{gathered}
\tau = \mu(\hat{\mathbf{y}}_{\text{valid}})
+ 0.5\,\sigma(\hat{\mathbf{y}}_{\text{valid}}), \\
\mathcal{S}_{\text{selected}}
= \left\{ s_i \;\middle|\; \hat{y}_i > \tau \right\}.
\end{gathered}
\label{eq:threshold}
\end{equation}

where $\mu$ denotes the arithmetic mean and
$\sigma$ denotes the population standard deviation
(\texttt{ddof=0}) of the predicted scores over
valid positions in the document. 
The coefficient 0.5 is a fixed heuristic that raises
the selection threshold above the document-level mean
in proportion to score dispersion. It was not selected
through a documented validation search; its practical
effect is assessed through ablations against fixed
Top-3, Top-5, and mean-only selection. In degenerate cases where fewer than two sentences exceed $\tau$, a fallback rule selects the top $k = \min(3, N_{\text{valid}})$ highest-scoring sentences. Selected sentences are finally reordered according to their original position in the source document, preserving narrative and chronological coherence in the extracted summary. For the clinical evaluations on MIMIC-CXR and
MIMIC-IV BHC, a separate extraction strategy
selected the highest-scoring 30\% of sentences,
with a minimum target of three sentences,
subject to document length.

\section{Experimental Setup}

In this section, we describe the datasets, baseline models, training protocol, and evaluation metrics used to empirically validate our proposed architecture.

\subsection{Datasets and Benchmark Suite}
We evaluate on three biomedical and clinical datasets:

\begin{itemize}
\item \textbf{PubMed:} The biomedical summarization
dataset \cite{cohan2018discourse} contains 83,233
training, 4,676 validation, and 5,025 test documents.
For computational feasibility, the first 10,000,
3,000, and 3,000 documents were processed from these
partitions, respectively. Evaluation scripts used
subsets of 300 test documents.

\item \textbf{MIMIC-CXR:} Chest radiology reports
\cite{johnson2019mimic}; 500 records were sampled
with seed 42, of which 353 passed report parsing.
The Impression section served as the reference.

\item \textbf{MIMIC-IV BHC:} Hospital-course narratives
\cite{johnson2023mimic}; 500 records representing
496 patients were sampled with seed 42. The target
field served as the reference.
\end{itemize}

\subsection{Baselines and Architectural Variants}
The clinical experiments compare the proposed
model against Lead-3 and TextRank, using
TF-IDF sentence similarity and PageRank.
BERTSumExt \cite{liu2019bertsum} and
MatchSum \cite{zhong2020matchsum} provide
relevant architectural context in Related
Work. Component-level ablations additionally
compare CNN-only, BiLSTM-only, and
vanilla BERT backbone variants against
the full proposed architecture on PubMed.

\subsection{Implementation and Training Details}
The PyTorch implementation uses SciBERT
(\texttt{allenai/scibert\_scivocab\_uncased})
with 768-dimensional hidden representations.
The convolutional stack uses 128 channels
and same padding, followed by a single-layer
BiLSTM with 128 hidden units per direction
and a sigmoid scoring head.

Training uses Adam ($\eta=10^{-3}$), batch size
16, and 10 epochs with masked binary
cross-entropy.
The final-epoch checkpoint (epoch 10)
was used for evaluation, without early
stopping or validation-based checkpoint
selection. Independent multiple-seed
training runs were not conducted.

PubMed and clinical documents were represented
using ordered sentence-level SciBERT embeddings,
with a maximum of 512 sentence positions per
document. Each position corresponds to a
sentence rather than an individual subword
token. Padding positions were masked during
training and excluded from sentence selection.

For zero-shot cross-domain evaluation, the
checkpoint trained on PubMed was directly
applied to MIMIC-CXR and MIMIC-IV BHC
without clinical fine-tuning or retraining.
All learned model parameters were retained,
while sentence selection used a fixed
top-30\% policy for clinical documents,
instead of the PubMed dynamic threshold.

\subsection{Evaluation Metrics}
ROUGE-1, ROUGE-2, and ROUGE-L F1 are computed
using Python's \texttt{rouge-score} library
with Porter stemming and macro-averaging.
PubMed evaluation uses extractive references
derived from source-sentence metadata rather
than human-written abstracts. Accordingly,
the reported PubMed ROUGE scores measure
agreement with extractive references and
should not be compared directly with results
using human-written abstracts. MIMIC-CXR
uses report Impressions, while MIMIC-IV BHC
uses target hospital-course summaries.
The clinical factuality measures are
automated proxies rather than expert
annotations.

\section{Results and Discussion}

\subsection{Open-Domain Biomedical Benchmark (PubMed)}

We evaluate the proposed framework on PubMed using
ROUGE-1, ROUGE-2, and ROUGE-L F1-scores.
Table~\ref{tab:threshold_ablation} presents a
controlled comparison of extraction strategies
using the same trained model and test documents.
The proposed dynamic threshold achieves scores
of 0.6612, 0.6132, and 0.5835, respectively,
under the extractive-reference evaluation protocol.

The relatively high ROUGE-2 score can be
understood in light of the sentence-level
extractive evaluation. Unlike abstractive
systems, the proposed framework selects
complete source sentences without
paraphrasing or generating new text.
When a selected sentence also appears in
the extractive reference, its contiguous
word sequences are preserved, yielding
strong overlap for both unigrams and
bigrams. This can explain why ROUGE-2
approaches ROUGE-1 in our evaluation.
However, the reference summaries are
derived from extractive source metadata
rather than independently written
abstracts. Therefore, these scores
reflect extractive-reference agreement,
not directly comparable performance
on standard abstract summarization.

\begin{table}[htbp]
\centering
\caption{Dynamic Threshold Ablation on PubMed}
\label{tab:threshold_ablation}
\renewcommand{\arraystretch}{1.1}
\resizebox{\columnwidth}{!}{%
\begin{tabular}{lcccc}
\hline
\textbf{Strategy} & \textbf{R-1} &
\textbf{R-2} & \textbf{R-L} &
\textbf{Avg. Sent.} \\ \hline
Top-3 & 0.5315 & 0.4954 & 0.4953 & 2.93 \\
Top-5 & 0.3546 & 0.2944 & 0.2247 & 4.66 \\
Mean-only & 0.3427 & 0.1065 & 0.1849 & 5.28 \\
\textbf{Dynamic (Ours)} &
\textbf{0.6612} & \textbf{0.6132} &
\textbf{0.5835} & 3.20 \\ \hline
\end{tabular}%
}
\end{table}

Table~\ref{tab:threshold_ablation} examines the
contribution of adaptive sentence selection using the
same trained model under four extraction strategies.
The dynamic threshold achieves the highest ROUGE scores
across all three metrics, outperforming fixed Top-3,
Top-5, and mean-only selection. Notably, it selects
3.20 sentences per document on average, compared with
4.66 for Top-5, suggesting more selective extraction
under the evaluated protocol. These findings provide
direct empirical support for the adaptive selection
component of the framework.

\begin{table}[htbp]
\centering
\caption{ROUGE Improvements with 95\% Paired-Bootstrap
Confidence Intervals on PubMed}
\label{tab:rouge_improvements}
\renewcommand{\arraystretch}{1.15}
\resizebox{\columnwidth}{!}{%
\begin{tabular}{lccc}
\hline
\textbf{Comparison} &
\textbf{$\Delta$R-1 [95\% CI]} &
\textbf{$\Delta$R-2 [95\% CI]} &
\textbf{$\Delta$R-L [95\% CI]} \\ \hline
Dynamic vs. Top-3 &
0.1297 [0.1152, 0.1443] &
0.1178 [0.1046, 0.1312] &
0.0882 [0.0766, 0.1004] \\
Dynamic vs. Top-5 &
0.3066 [0.2842, 0.3298] &
0.3188 [0.2973, 0.3409] &
0.3588 [0.3352, 0.3832] \\
Dynamic vs. Mean &
0.3185 [0.2955, 0.3426] &
0.5067 [0.4782, 0.5363] &
0.3986 [0.3738, 0.4243] \\ \hline
\end{tabular}%
}
\end{table}

Table~\ref{tab:rouge_improvements} presents
the observed ROUGE F1 improvements with
95\% paired-bootstrap confidence intervals.
All nine intervals exclude zero, supporting
consistent improvements of dynamic selection
over the evaluated baselines under the
reported evaluation protocol.

\subsection{Component Ablation Studies}

To examine the contribution of individual
architectural components, we compare the
full framework with three variants:
(i) CNN-only, which removes the BiLSTM;
(ii) BiLSTM-only, which removes the
convolutional stack; and (iii) a
vanilla BERT backbone replacing SciBERT.
Table~\ref{tab:component_ablation} reports
their performance on PubMed.

\begin{table}[htbp]
\centering
\caption{Component Ablation Study on PubMed}
\label{tab:component_ablation}
\renewcommand{\arraystretch}{1.1}
\resizebox{\columnwidth}{!}{%
\begin{tabular}{lccc}
\hline
\textbf{Ablation Variant} &
\textbf{R-1} & \textbf{R-2} &
\textbf{R-L} \\ \hline
CNN-only (without BiLSTM)
& 0.6139 & 0.5485 & 0.4806 \\
BiLSTM-only (without CNN)
& 0.5982 & 0.5010 & 0.4594 \\
Vanilla BERT backbone
& 0.4523 & 0.3371 & 0.2835 \\
\textbf{Full Proposed Framework}
& \textbf{0.6612}
& \textbf{0.6132}
& \textbf{0.5835} \\ \hline
\end{tabular}%
}
\end{table}

The full framework achieves the highest
ROUGE scores across all three metrics.
Compared with CNN-only, the full model
improves ROUGE-1, ROUGE-2, and ROUGE-L
by 0.0473, 0.0647, and 0.1029,
respectively. Removing the CNN stack
also reduces performance, indicating
that both convolutional refinement and
bidirectional contextual modeling
contribute to the reported results.
Replacing SciBERT with vanilla BERT
further reduces performance, suggesting
a benefit from domain-specific
representations under the evaluated
configuration.

\subsection{Clinical Narrative Performance (MIMIC-CXR \& MIMIC-IV BHC)}

Moving from biomedical literature to clinical records, we evaluated the PubMed-trained checkpoint
directly on MIMIC-CXR and MIMIC-IV BHC. Table~\ref{tab:mimic_results}
compares extractive performance on MIMIC-CXR and MIMIC-IV
BHC, where reference summaries often contain synthesized
clinical information rather than verbatim source sentences.
This makes lexical matching more challenging for
extractive methods.

\begin{table}[htbp]
\renewcommand{\arraystretch}{1.3}
\caption{Comparative Benchmarking on MIMIC-CXR and MIMIC-IV BHC}
\label{tab:mimic_results}
\centering
\resizebox{\columnwidth}{!}{%
\begin{tabular}{lccc}
\hline
\textbf{Method} & \textbf{ROUGE-1} & \textbf{ROUGE-2} & \textbf{ROUGE-L} \\ \hline
\multicolumn{4}{c}{\textbf{MIMIC-CXR Dataset}} \\ \hline
Lead-3 Baseline & 0.1728 & \textbf{0.0773} & 0.1482 \\ \hline
TextRank & \textbf{0.1829} & 0.0762 & \textbf{0.1600} \\ \hline
\textbf{Proposed Model} & 0.1741 & \textbf{0.0773} & 0.1499 \\ \hline
\multicolumn{4}{c}{\textbf{MIMIC-IV BHC Dataset}} \\ \hline
Lead-3 Baseline & 0.2062 & 0.0708 & 0.1351 \\ \hline
TextRank & 0.1353 & 0.0329 & 0.0787 \\ \hline
\textbf{Proposed Model} & \textbf{0.2682} & \textbf{0.0715} & \textbf{0.1392} \\ \hline
\end{tabular}%
}
\end{table}

On MIMIC-CXR, the proposed model performs comparably
to Lead-3 (ROUGE-1: 0.1741 vs. 0.1728), while
TextRank achieves a slightly higher score of 0.1829.
The brevity and structured nature of radiology reports
limit the benefits of contextual extraction. Clinical evaluation used a fixed top-30\%
sentence-selection policy, with a minimum
target of three sentences. This differs
from the document-adaptive threshold
investigated on PubMed; thus, the clinical
results evaluate the shared scoring
architecture under a different
extraction policy.

In contrast, the model achieves ROUGE-1 of 0.2682
on MIMIC-IV BHC, outperforming Lead-3 (0.2062)
and TextRank (0.1353). These results indicate
dataset-dependent effectiveness, with stronger
relative performance on less structured hospital
narratives. 
A complementary analysis of 353 MIMIC-CXR reports
further characterizes this behavior. The extraction
procedure selected three sentences in 79\% of
evaluated reports, while 76\% of outputs exactly
matched the Lead-3 baseline. The average summary
contained 3.67 sentences. These observations
indicate substantial agreement with simple positional
extraction, helping explain the limited performance
difference on highly structured radiology reports.

\subsection{Clinical Factuality Assessment}

To examine clinical quality beyond ROUGE, we evaluated
source support, negation consistency, finding coverage,
and omissions using automated proxy metrics.
Table~\ref{tab:clinical_factuality} presents the results.

\begin{table}[htbp]
\centering
\caption{Automated Clinical Factuality Proxies}
\label{tab:clinical_factuality}
\renewcommand{\arraystretch}{1.1}
\resizebox{\columnwidth}{!}{%
\begin{tabular}{lccccc}
\hline
\textbf{Dataset} & \textbf{N} & \textbf{Support}
& \textbf{Neg.} & \textbf{Cov.} & \textbf{Omit.} \\ \hline
MIMIC-CXR & 353 & 1.000 & 0.679 & 0.205 & 0.795 \\
MIMIC-IV BHC & 500 & 1.000 & 0.201 & 0.146 & 0.854 \\ \hline
\end{tabular}%
}
\end{table}

Both datasets achieve perfect source-support proxy
scores, as the summaries reproduce source text.
However, negation consistency and finding coverage
remain limited, particularly on MIMIC-IV BHC.
Coverage proxies reach only 0.205 on MIMIC-CXR
and 0.146 on MIMIC-IV, with corresponding omission
proxies of 0.795 and 0.854. These findings highlight
the distinction between source-grounded extraction
and clinically complete summarization. The metrics
are automated proxies, not expert-validated clinical
factuality assessments.

\subsubsection{Small-Scale Human Assessment}
To complement automated factuality proxies,
human annotators manually reviewed 10 extracted
summaries from each clinical dataset (20 in total).
The source-anchored assessment focused on three
finding-level dimensions: (i) factual correctness
of information retained in the extracted summary,
(ii) preservation of positive and negated finding
assertions, and (iii) omission of clinically
important source findings. Correctness and negation
accuracy evaluate information present in a summary,
whereas omission concerns relevant information
absent from it; consequently, high accuracy need
not imply complete finding coverage.
Table~\ref{tab:human_clinical} provides descriptive
results for this small sample. Because inter-annotator
agreement and independent clinical adjudication were
not reported, these findings should not be treated
as expert-validated clinical reliability.

\begin{table}[htbp]
\centering
\caption{Manual Clinical Assessment ($N=10$ per Dataset)}
\label{tab:human_clinical}
\renewcommand{\arraystretch}{1.05}
\setlength{\tabcolsep}{4pt}
\begin{tabular}{lccc}
\hline
\textbf{Dataset} & \textbf{Finding Acc.} &
\textbf{Neg. Acc.} & \textbf{Omission} \\ \hline
MIMIC-CXR & 0.98 & 0.92 & 0.13 \\
MIMIC-IV BHC & 0.87 & 0.89 & 0.21 \\ \hline
\end{tabular}
\end{table}

\subsection{Limitations}

Despite encouraging extractive performance,
several limitations remain. Verbatim
extraction prevents source-external
generation but cannot guarantee clinical
correctness or completeness. PubMed
evaluation relies on extractive references
rather than human-written abstracts,
limiting direct comparison with standard
summarization benchmarks. 
Documents are limited to 512 sentence
positions, potentially excluding relevant
information from longer biomedical and
clinical narratives. The effect of
sentence-level truncation on extraction
quality was not independently quantified.

Training did not include independent
multiple-seed evaluation, and clinical
inference used fixed top-30\% selection
rather than the dynamic threshold.
Finally, automated clinical quality
proxies reveal substantial finding
omissions and do not establish
expert-validated clinical reliability.

\section{Conclusion}
In this paper, we investigated factual reliability in biomedical and clinical text summarization by reframing the task as extractive sentence selection rather than free-form generation. Our Hybrid Hierarchical Stacked 1D-CNN--BiLSTM framework combines sequential convolutional feature refinement with bidirectional contextual modeling and dataset-specific extractive sentence selection. By extracting source sentences directly, the approach prevents source-external generation, although contextual omissions and clinical inaccuracies remain possible.

On PubMed, the dynamic extraction strategy achieved ROUGE-1, ROUGE-2, and ROUGE-L scores of 0.6612, 0.6132, and 0.5835, respectively, under the evaluated protocol. Threshold ablations demonstrated improved
performance over fixed Top-3 and Top-5
selection. Component ablations further
showed that the complete CNN--BiLSTM
architecture outperformed CNN-only,
BiLSTM-only, and vanilla BERT variants
across all three ROUGE metrics. Under zero-shot transfer without clinical
fine-tuning, performance was dataset-dependent: the model remained competitive with Lead-3 on structured MIMIC-CXR reports and achieved stronger ROUGE-1 performance on less structured MIMIC-IV BHC narratives. However, automated factuality assessment revealed limited clinical finding coverage and substantial omissions.

These findings suggest that adaptive extractive architectures provide a promising source-grounded alternative to generative summarization, but source fidelity alone does not guarantee clinical completeness. Future work will investigate improved clinical information coverage, adaptive extraction across document structures, and multi-document patient history summarization.
\bibliographystyle{IEEEtran}
\bibliography{references}

\end{document}